\documentclass[runningheads]{llncs}
\usepackage{graphicx}
\usepackage{booktabs}
\usepackage{multirow}
\usepackage{array}
\usepackage{tabularx}
\usepackage{amsmath}
\usepackage{microtype}
\usepackage[hidelinks]{hyperref}
\begin{document}
\title{GBU-Palm: A Multimodal Video Dataset and Benchmark for Palm Presentation Attack Detection}
\titlerunning{GBU-Palm: Multimodal Video Palm PAD Benchmark}

\makeatletter
\def\@fnsymbol#1{\ensuremath{%
  \ifcase#1%
  \or\star%
  \or\star\star%
  \or\star\star\star%
  \or\ddagger%
  \or\mathchar"278%
  \or\mathchar"27B%
  \or\|%
  \or **%
  \or\dagger\dagger%
  \or\ddagger\ddagger%
  \else\@ctrerr%
  \fi}}
\makeatother

\author{
Yingjie Ma\inst{1,2}
\and
Zitong Yu\inst{2,3,4}\thanks{Corresponding authors}
\and
Wei Jia\inst{5}
\and
Ajay Kumar\inst{6}
\and
Linlin Shen\inst{1,4}\protect\footnotemark[1]
}
\authorrunning{Y.~Ma et al.}

\institute{
\begin{tabular}{@{}c@{}}
\inst{1} Shenzhen University \qquad
\inst{2} Great Bay University \\[-1pt]
\inst{3} Dongguan Key Laboratory for Intelligence and Information Technology \\[-1pt]
\inst{4} Guangdong Provincial Key Laboratory of Intelligent Information Processing,
Shenzhen University \\[-1pt]
\inst{5} Hefei University of Technology \qquad
\inst{6} The Hong Kong Polytechnic University
\end{tabular}
}


\maketitle
\begin{abstract}
Existing palm presentation attack detection (PAD) datasets are often limited by static imagery, restricted acquisition conditions, or insufficient multimodal video data, hindering systematic evaluation across environments, modalities, and attack types. We present GBU-Palm, a large-scale multimodal video dataset and benchmark containing 21,326 videos from 105 subjects and 210 palms across six acquisition environments, including bona fide, Print, and Replay presentations, with 6,310 synchronized RGB-NIR samples. We construct leakage-controlled protocols that separate palm identity and attack lineage and benchmark four representative video architectures under environment-matched and held-out-environment settings. Results reveal substantial architecture-dependent degradation under environmental shift and show that RGB-NIR fusion does not consistently outperform RGB-only input. We further analyze model behavior through true accept (TA), true reject (TR), false accept (FA), and false reject (FR) decomposition, spectral masking, temporal-order intervention, and frozen-backbone NIR probing, revealing distinct failure patterns and evidence utilization across architectures. GBU-Palm provides a unified and challenging benchmark for developing and evaluating robust multimodal palm PAD methods under cross-environment conditions. The dataset will be released soon.
\keywords{Palm presentation attack detection; Multimodal; RGB-NIR}
\end{abstract}

\section{Introduction}

Palm biometrics provide a convenient and contactless means of authentication, but printed or screen-replayed palm content can be presented directly to the sensor, creating practical presentation attacks. Presentation attack detection (PAD) is therefore essential for reliable deployment. Early studies demonstrated vulnerabilities of palmprint and palm-vein systems to Print and Display/Replay attacks and introduced representative resources such as PALMspoof and VERA Spoofing PalmVein~\cite{bhilare2018vulnerability,tome2015vulnerability}. More recent work has explored cross-domain adaptation, domain generalization, frequency cues, and larger NIR palm-vein PAD data~\cite{yao2023palmprint,liu2025domain,liu2025hfsra,yan2026pvasd}. Nevertheless, representative palm PAD resources remain largely based on static imagery, single-spectrum sensing, or limited acquisition conditions, making it difficult to systematically evaluate video PAD generalization across attack types, sensing modalities, and environments.

Video and multimodal sensing introduce two important dimensions for palm PAD. Bona fide palms, printed media, and replay displays differ not only in appearance, but also in motion consistency, surface stability, reflections, and display dynamics. Video PAD in other biometric domains has demonstrated the value of spatiotemporal information~\cite{liu2018deepfas,yang2019modeldata}. Meanwhile, RGB and near-infrared (NIR) capture different responses from skin, printed materials, and electronic displays, while prior multimodal PAD studies show that additional sensing channels do not necessarily improve performance~\cite{zhang2019casiasurf,george2021crossmodal}. These factors may also change across acquisition environments, and palm PAD has already exhibited substantial cross-device and cross-domain degradation~\cite{yao2023palmprint,liu2025domain}. Reliable comparison of video, multimodal, and cross-environment generalization therefore requires these factors to be evaluated jointly within a controlled data design.

\begin{table}[t]
\centering
\caption{Representative palm PAD resources. Scale follows the evaluation unit reported by each source and is not directly comparable across rows.}
\label{tab:dataset_comparison}
\small
\setlength{\tabcolsep}{3.0pt}
\renewcommand{\arraystretch}{1.03}
\begin{tabular}{@{}l c c r c c@{}}
\toprule
\textbf{Dataset} & \textbf{Access} & \textbf{Unit} & \textbf{Scale} & \textbf{Modality} & \textbf{Attack} \\
\midrule
VERA~\cite{tome2015vulnerability}& Request & Image & 2,000 & NIR & Print \\
PALMspoof~\cite{bhilare2018vulnerability}& Private & Image & -- & RGB & Print/display \\
XJTU-PalmReplay~\cite{yao2023palmprint}& Request & Image & 96,000 & RGB & Replay \\
PVASD~\cite{yan2026pvasd}& Public & Image & 1,187,519 & NIR & 2D/3D \\
\textbf{GBU-Palm}& \textbf{Public} & \textbf{Video} & \textbf{21,326} & \textbf{RGB/NIR} & \textbf{Print/replay} \\
\bottomrule
\end{tabular}
\end{table}

However, as summarized in Table~\ref{tab:dataset_comparison}, existing palm PAD datasets do not jointly provide large-scale native video, synchronized RGB-NIR observations, multiple acquisition environments, and controlled presentation-attack provenance within a unified benchmark. Consequently, several practically important questions remain difficult to study reproducibly, including how strongly performance degrades under environmental shift, whether RGB and NIR provide complementary information for different architectures, whether video models actually depend on temporal ordering, and whether different attack types and error categories exhibit the same failure trends. Addressing these questions requires not only larger-scale data, but also a standardized benchmark that controls identity, attack origin, modality, temporal structure, and environment.

To address this gap, we introduce \textbf{GBU-Palm}, a large-scale multimodal video dataset and benchmark for palm presentation attack detection. GBU-Palm contains 21,326 videos from 105 subjects and 210 palms across six acquisition environments, covering bona fide, Print, and Replay presentations, with 6,310 synchronized RGB-NIR samples. We construct evaluation protocols with disjoint palm identities and attack lineages and benchmark representative video architectures under RGB, NIR, and RGB-NIR settings in both environment-matched (In-Env) and held-out-environment (Cross-Env) conditions. Beyond the overall benchmark, we further study model failure modes and spectral-temporal information utilization through TA/TR/FR/FA decision-outcome decomposition, spatiotemporal evidence analysis, spectral masking, temporal-order intervention, and frozen-backbone NIR probing. Our contributions are threefold:
\begin{itemize}

\item We introduce \textbf{GBU-Palm}, a large-scale multimodal video dataset for palm PAD containing 21,326 videos from 105 subjects and 210 palms across six acquisition environments, including bona fide, Print, and Replay presentations, 6,310 synchronized RGB-NIR samples, and attack-lineage annotations.

\item We establish standardized, leakage-controlled benchmark protocols with identity- and attack-lineage-disjoint splits, and systematically evaluate representative video architectures under RGB, NIR, and RGB-NIR inputs in both In-Env and Cross-Env conditions.

\item Beyond the overall benchmark, we further analyze model error patterns and spatiotemporal and spectral information utilization. The results reveal architecture-dependent cross-environment degradation, non-uniform benefits from RGB-NIR fusion, and substantially different dependence on temporal ordering and NIR information across video models.

\end{itemize}

\section{GBU-Palm Dataset}

\begin{table}[t]
\centering
\caption{Presentation-attack factors retained in GBU-Palm metadata.}
\vspace{-0.8em}
\label{tab:attack_factors}
\small
\setlength{\tabcolsep}{3.5pt}
\begin{tabularx}{\linewidth}{@{}l l X@{}}
\toprule
\textbf{Attack} & \textbf{Factor} & \textbf{Values} \\
\midrule
Print & Spectrum & RGB color, grayscale, NIR \\
Print & Material & A4, photo, matte, pearl, coated paper \\
Print & Crop & Full, local, hand-shape, partial real-palm exposure \\
Replay & Spectrum & RGB, NIR \\
Replay & Interface & Full-screen, UI borders \\
\bottomrule
\end{tabularx}
\end{table}

\noindent \textbf{Acquisition and Attack Scenarios.} The released GBU-Palm dataset contains 21,326 videos from 105 subjects and 210 palms acquired under six illumination environments (E1--E6) using consumer phones, tablets, laptops, and a synchronized RGB-NIR acquisition system. E1 represents uniform indoor normal illumination, including artificial, mixed artificial-natural, and indirect natural light sampled at $1{:}1{:}1\pm10\%$; E2 represents indoor low illumination, with weak artificial and weak artificial-natural light sampled at $2{:}1\pm10\%$; E3 represents indoor directional illumination, covering natural or artificial side and back lighting at $1{:}1\pm10\%$; E4 represents uniform outdoor indirect natural light under cloudy, overcast, or shaded conditions; E5 represents uniform outdoor indirect natural light under explicit shelter; and E6 represents outdoor directional natural light with side and back lighting sampled at $1{:}1\pm10\%$. All environments follow a common data organization and ROI convention, enabling acquisition conditions to be compared under a consistent representation. The dataset contains three presentation classes. \emph{Bona fide} videos record genuine palms. \emph{Print} attacks recapture printed palm media generated from genuine source samples, whereas \emph{Replay} attacks present palm content through electronic displays and recapture the resulting presentation. For each attack sample, the metadata retains its source bona fide session, attack generation, and material identity, forming an \emph{attack lineage}. Beyond the attack labels, GBU-Palm retains interpretable generation factors including printing spectrum, physical medium, crop strategy, replay spectrum, and playback interface, enabling analyses beyond the coarse Print/Replay taxonomy.

\begin{figure}[t]
\centering
\includegraphics[width=0.8\linewidth]{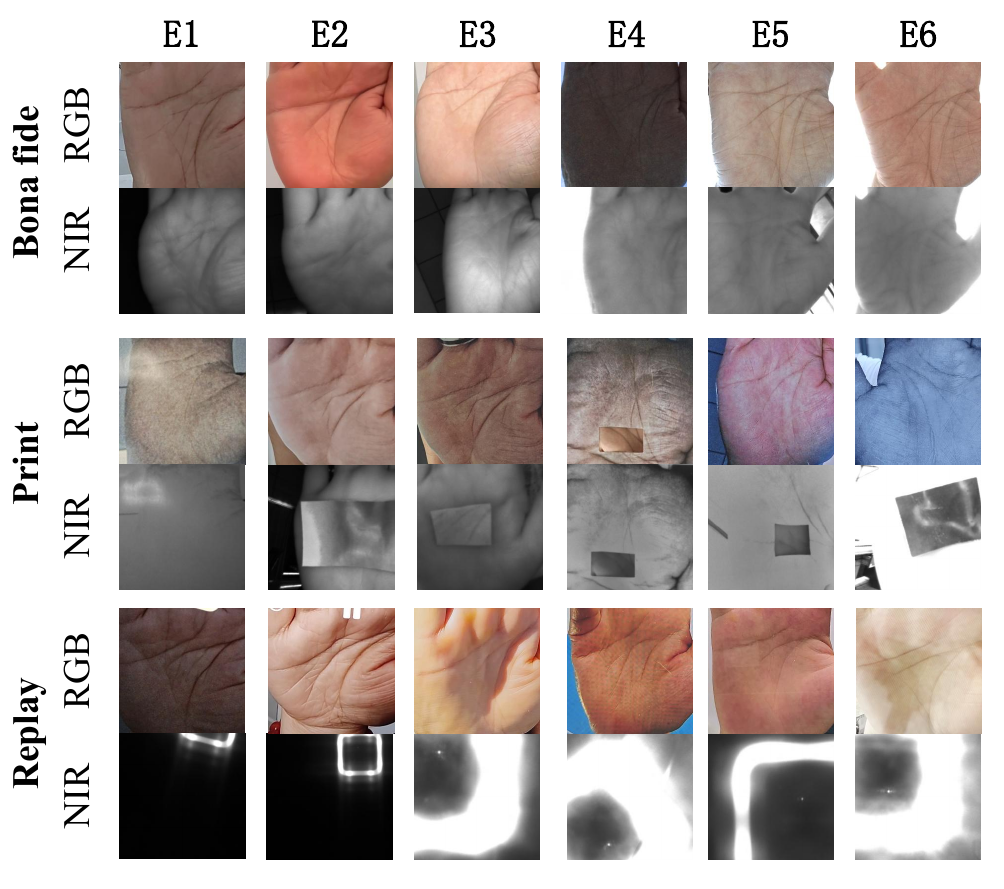}
\vspace{-2.0em}
\caption{Representative GBU-Palm samples across six acquisition environments
(E1--E6) and three presentation classes. RGB and NIR rows show synchronized
observations of the same physical presentations, illustrating variation across
environment, attack type, and sensing spectrum.}
\label{fig:examples}
\end{figure}

\noindent \textbf{Native Video and Multimodal Construction.} Each GBU-Palm sample preserves a continuous 16-s physical observation window together with its native temporal ordering and frame timestamps, with palm observations stored as canonical $256\times256$ ROIs. Unlike video sequences constructed from static imagery, these native videos retain motion, reflection, and presentation dynamics produced during physical acquisition. For synchronized RGB-NIR samples, both modalities record the same physical presentation over the same observation window. RGB serves as the common spatial-geometry reference and is mapped to NIR through the fixed acquisition-system geometry, preserving consistent ROI definitions and spatial correspondence across modalities, while both streams retain their native frame timestamps.

\noindent \textbf{Dataset Statistics.} The released set contains 11,108 bona fide videos, 5,584 Print attacks, and 4,634 Replay attacks. Among the 21,326 samples, 15,016 are RGB-only and 6,310 provide synchronized RGB-NIR observations, including 4,471 bona fide, 1,407 Print, and 432 Replay samples. In total, the canonical media store contains 27,636 modality sequences and 5,293,639 frames; unobserved NIR data are not synthesized. All 105 subjects have recorded age and sex metadata, comprising 51 male and 54 female participants. Ages range from 18 to 70 years, with a mean of $44.1\pm16.6$ years and a median of 45. The 18--30, 31--45, 46--60, and 61+ age groups contain 26, 27, 26, and 26 subjects, respectively, resulting in near-balanced distributions across sex and the predefined age groups.

\noindent \textbf{Benchmark Design.} The GBU-Palm benchmark formulates palm PAD as binary discrimination between bona fide and attack presentations. For each video, models select 32 unique real-frame positions, and the canonical $256\times256$ ROI is transformed to a $112\times112$ model input. No temporal interpolation, frame duplication, or fabricated frames are introduced, so all frames presented to the model originate from physically captured video content. RGB, NIR, and RGB-NIR settings follow common training and evaluation rules. We define two complementary evaluation protocols. P1 evaluates environment-matched (\emph{In-Env}) performance: subjects are disjoint across training, validation, and test splits, while all six acquisition environments are represented in each partition. P2 evaluates cross-environment (\emph{Cross-Env}) generalization: E2, E4, E5, and E6 are used for training and validation, whereas E1 and E3 are completely withheld from training and used exclusively for testing. Both protocols enforce disjoint subjects, palm identities, and attack lineages between training and evaluation partitions. Table~\ref{tab:protocol} reports the resulting split composition.

\begin{table}[t]
\centering
\caption{Split composition of the GBU-Palm P1 (In-Env) and P2 (Cross-Env) protocols. RGB+NIR denotes samples with synchronized paired observations.}
\label{tab:protocol}
\vspace{-0.8em}
\small
\renewcommand{\arraystretch}{1.04}
\setlength{\tabcolsep}{3.7pt}
\begin{tabular}{@{}llrrrrr@{}}
\toprule
\textbf{Protocol} & \textbf{Split} & \textbf{Total} & \textbf{Bona fide} & \textbf{Print} & \textbf{Replay} & \textbf{RGB+NIR} \\
\midrule
P1 & Train & 12,152 & 7,433 & 2,049 & 2,670 & 3,740 \\
   & Val   & 3,920  & 1,649 & 1,294 & 977   & 1,115 \\
   & Test  & 3,422  & 1,641 & 902   & 879   & 1,031 \\
\midrule
P2 & Train & 4,827  & 3,455 & 445   & 927   & 1,482 \\
   & Val   & 1,975  & 1,040 & 618   & 317   & 529 \\
   & Test  & 2,174  & 1,471 & 279   & 424   & 836 \\
\bottomrule
\end{tabular}
\end{table}

\section{Benchmark Results and Evidence Analysis}

\noindent \textbf{Experimental Setup.} We benchmark four representative video backbones with different temporal modeling inductive biases: R(2+1)D-18~\cite{tran2018r2plus1d}, a factorized ViViT baseline adapted from ViViT~\cite{arnab2021vivit}, Video Swin-T~\cite{liu2022videoswin}, and MViT-V2-S~\cite{li2022mvitv2}. R(2+1)D represents factorized spatiotemporal convolution, ViViT represents transformer-based video modeling, while Video Swin-T and MViT-V2-S cover hierarchical and multi-scale spatiotemporal transformer architectures. This diversity enables analysis of how different video models exploit temporal and spectral evidence in palm PAD. Each model is trained independently under P1 and P2 protocols. RGB-only samples provide RGB streams, while synchronized RGB-NIR samples provide both modalities. The two streams share the same backbone and their available embeddings are averaged before the classification head. This design avoids introducing an additional fusion module and keeps the comparison focused on evidence utilization rather than fusion architecture optimization. All models use identical optimization settings. Inputs consist of 32 real-frame positions with spatial resolution of $112\times112$. Training uses AdamW with learning rate $2\times10^{-4}$, weight decay $10^{-4}$, a maximum of 100 epochs, and patience of 15. Checkpoints are selected according to validation AUC, and test data are never used for checkpoint selection or threshold determination. We report AUC to measure threshold-independent ranking performance and HTER, defined as the average of false acceptance rate (FAR) and false rejection rate (FRR), to evaluate PAD errors under a fixed decision threshold.

\noindent\textbf{Overall benchmark performance.}
Table~\ref{tab:main} reports results on the complete test splits. Under P1 environment-matched evaluation, performance already varies substantially across architectures: R(2+1)D and MViT reach approximately 97.4\% AUC, whereas Video Swin-T performs considerably worse. Under P2 Cross-Env evaluation, AUC decreases for all four architectures, but by markedly different amounts: ViViT loses 11.28 points, compared with 2.87 for R(2+1)D, 1.75 for Video Swin-T, and 4.03 for MViT. Cross-environment generalization is therefore strongly architecture-dependent rather than a uniform increase in task difficulty.

\begin{table}[t]
\centering
\caption{Overall video-based PAD results on the complete GBU-Palm test splits. AUC and HTER are reported in percentage; HTER uses the operating threshold fixed on validation.}
\label{tab:main}
\vspace{-0.8em}
\small
\renewcommand{\arraystretch}{1.07}
\setlength{\tabcolsep}{5.5pt}
\resizebox{0.82\textwidth}{!}{%
\begin{tabular}{l|cc|cc}
\toprule
\multirow{2}{*}{\textbf{Method}}
& \multicolumn{2}{c|}{\textbf{P1: In-Env}}
& \multicolumn{2}{c}{\textbf{P2: Cross-Env}} \\
\cmidrule(lr){2-3}\cmidrule(lr){4-5}
& \textbf{AUC$\uparrow$ (\%)} & \textbf{HTER$\downarrow$ (\%)}
& \textbf{AUC$\uparrow$ (\%)} & \textbf{HTER$\downarrow$ (\%)} \\
\midrule
R(2+1)D-18~\cite{tran2018r2plus1d} & \textbf{97.48} & \textbf{8.39} & \textbf{94.61} & \textbf{13.05} \\
ViViT~\cite{arnab2021vivit} & 93.49 & 14.55 & 82.21 & 25.73 \\
Video Swin-T~\cite{liu2022videoswin} & 72.95 & 32.89 & 71.20 & 36.36 \\
MViT-V2-S~\cite{li2022mvitv2} & 97.33 & \textbf{8.39} & 93.30 & 17.00 \\
\bottomrule
\end{tabular}%
}
\end{table}

\noindent\textbf{Environment-specific generalization.}
As shown in Table~\ref{tab:environment_generalization}, environment sensitivity is strongly architecture-dependent. Across the six P1 environments, Video Swin-T spans 10.0 AUC points and ViViT 4.6 points, whereas R(2+1)D and MViT vary by only 2.2 and 1.7 points, respectively. The best and worst environments also differ across architectures, indicating no universal environment-difficulty ordering. Within P2, E3 yields lower AUC than E1 for all four models, with the largest gap of 2.7 points observed for MViT.

\begin{table}[t]
\centering
\caption{Environment-specific PAD performance (AUC, \%). P1 Range denotes the
maximum--minimum AUC across the six P1 environments. P2 values compare the two
held-out environments within the same protocol.}
\label{tab:environment_generalization}
\vspace{-0.8em}
\small
\setlength{\tabcolsep}{3.5pt}
\resizebox{0.8\columnwidth}{!}{%
\begin{tabular}{lccccc}
\toprule
\textbf{Method} & \textbf{P1 Best} & \textbf{P1 Worst} &
\textbf{Range} & \textbf{P2 E1} & \textbf{P2 E3} \\
\midrule
R(2+1)D-18 & E1 / 98.7 & E6 / 96.5 & 2.2 & 95.2 & 94.1 \\
ViViT      & E2 / 95.8 & E5 / 91.2 & 4.6 & 82.6 & 81.8 \\
Video Swin-T & E6 / 77.1 & E4 / 67.1 & 10.0 & 71.7 & 70.6 \\
MViT-V2-S  & E1 / 98.3 & E6 / 96.7 & 1.7 & 94.6 & 92.0 \\
\bottomrule
\end{tabular}%
}
\end{table}

\begin{figure}[t]
\centering
\includegraphics[width=0.92\linewidth]{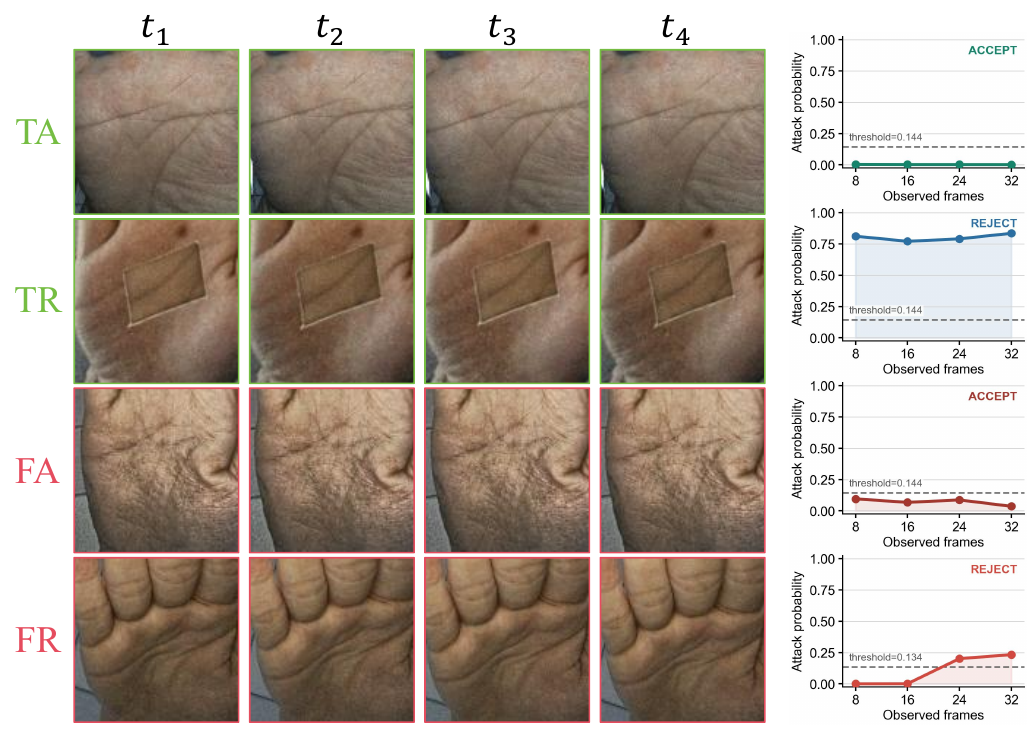}
\vspace{-1.8em}
\caption{Decision-conditioned examples from GBU-Palm. TA, TR, FA, and FR cases show representative temporal observations ($t_1$--$t_4$) and corresponding attack-probability trajectories. The visualization highlights different decision behaviors without making unsupported pixel-level attribution claims.}
\label{fig:decision_conditioned}
\end{figure}

\noindent\textbf{Decision-conditioned error structure.}
To expose failure directions hidden by aggregate AUC, we decompose PAD decisions into true accept (TA), true reject (TR), false accept (FA), and false reject (FR). TA denotes correctly accepted bona fide samples and TR correctly rejected attacks; FA denotes attacks incorrectly accepted as bona fide and directly reflects security risk, whereas FR denotes bona fide samples incorrectly rejected and reflects usability cost. Figure~\ref{fig:decision_conditioned} shows representative temporal observations and attack-probability trajectories for the four outcomes, illustrating that similar aggregate performance changes can correspond to fundamentally different decision failures. This asymmetry is pronounced under environmental shift. MViT FA rises from 7.52\% on P1 to 26.32\% on P2, while FR decreases from 9.26\% to 7.68\%. R(2+1)D FA increases from 9.77\% to 18.21\%, whereas FR remains relatively stable. ViViT, in contrast, increases in both FA and FR. Environmental shift can therefore primarily increase attack-acceptance risk for some architectures while simultaneously affecting both security and usability for others, behavior that is not captured by the magnitude of AUC degradation alone.

\noindent\textbf{Attack-family vulnerability.}
Further decomposing FA by attack family shows that Replay generally presents a higher attack-acceptance risk than Print across model--protocol combinations. For example, under MViT P2, Print FA is 11.47\%, whereas Replay FA reaches 36.08\%. Cross-environment vulnerability therefore depends not only on model architecture, but also on the physical presentation mechanism.

\noindent\textbf{Spectral Evidence Analysis.}
Synchronized RGB-NIR samples enable spectral comparisons under the same physical presentation conditions. This analysis investigates whether additional spectral information can be effectively utilized by different architectures rather than assuming that multimodal input is always beneficial. Table~\ref{tab:evidence} shows that RGB-NIR benefits are strongly architecture-dependent. R(2+1)D improves from 4.33\% HTER with RGB input to 3.02\% with RGB+NIR on P1, and from 7.97\% to 4.08\% on P2. ViViT also benefits from additional NIR information. However, Swin increases from 19.74\% HTER with RGB to 22.98\% with RGB+NIR on P1, indicating that additional modalities do not automatically translate into useful discriminative evidence. The Replay breakdown further shows how spectral changes affect False Accept errors. Since FA represents attacks incorrectly accepted as bona fide, Replay FA directly reflects security impact under different spectral inputs. On P1, NIR-only Replay FA reaches 67\%, 93\%, and 94\% for R(2+1)D, ViViT, and Swin, respectively, while RGB+NIR reduces them to 6\%, 0\%, and 38\%. MViT maintains 0\% FA under both conditions. These results indicate that RGB and NIR are not simply additive information sources; their effectiveness depends on whether a model can learn and exploit complementary cross-modal cues.

\begin{table}[t]
\centering
\caption{
Controlled spectral and temporal evidence.
(a) Paired-test AUC/HTER (\%) from the same multimodal-trained checkpoint
and identical sample identities.
(b) Full-test AUC under normal, shuffled, and reversed temporal order;
shuffled results report mean $\pm$ std over five fixed permutations.
All temporal interventions preserve the same set of observed frames.
}
\label{tab:evidence}
\vspace{-0.8em}
\small
\renewcommand{\arraystretch}{1.04}
\textbf{(a) Spectral observation: AUC / HTER}\par\vspace{2pt}
\resizebox{0.94\textwidth}{!}{%
\begin{tabular}{l|ccc|ccc}
\toprule
\multirow{2}{*}{\textbf{Method}} & \multicolumn{3}{c|}{\textbf{P1: In-Env}} & \multicolumn{3}{c}{\textbf{P2: Cross-Env}} \\
\cmidrule(lr){2-4}\cmidrule(lr){5-7}
& \textbf{RGB} & \textbf{NIR} & \textbf{RGB+NIR} & \textbf{RGB} & \textbf{NIR} & \textbf{RGB+NIR} \\
\midrule
R(2+1)D & 99.45/4.33 & 84.09/25.03 & 99.66/3.02 & 96.94/7.97 & 91.54/16.19 & 98.65/4.08 \\
ViViT & 95.61/9.13 & 67.71/44.54 & 98.82/4.82 & 83.75/21.43 & 67.14/45.41 & 87.15/19.14 \\
Swin-T & 87.42/19.74 & 30.69/65.64 & 83.67/22.98 & 78.84/25.33 & 77.97/20.52 & 82.39/21.92 \\
MViTv2-S & 97.25/6.66 & 99.66/2.61 & 99.74/1.77 & 97.20/6.77 & 99.07/5.47 & 99.38/5.58 \\
\bottomrule
\end{tabular}%
}
\par\vspace{5pt}
\textbf{(b) Temporal-order intervention: AUC}\par\vspace{2pt}
\resizebox{0.86\textwidth}{!}{%
\begin{tabular}{l|ccc|ccc}
\toprule
\multirow{2}{*}{\textbf{Method}}
& \multicolumn{3}{c|}{\textbf{P1: In-Env}}
& \multicolumn{3}{c}{\textbf{P2: Cross-Env}} \\
\cmidrule(lr){2-4}\cmidrule(lr){5-7}
& \textbf{Normal} & \textbf{Shuffled} & \textbf{Reversed}
& \textbf{Normal} & \textbf{Shuffled} & \textbf{Reversed} \\
\midrule
R(2+1)D
& 97.48 & 93.34$\pm$0.39 & 93.26
& 94.61 & 89.58$\pm$0.24 & 90.98 \\

ViViT
& 93.49 & 93.49$\pm$0.00 & 93.49
& 82.21 & 82.21$\pm$0.00 & 82.21 \\

Swin-T
& 72.95 & 73.31$\pm$0.35 & 72.90
& 71.20 & 72.21$\pm$0.12 & 71.17 \\

MViTv2-S
& 97.33 & 95.95$\pm$0.13 & 94.69
& 93.30 & 90.90$\pm$0.14 & 91.39 \\
\bottomrule
\end{tabular}%

}
\end{table}

\noindent\textbf{Temporal Evidence Analysis.}
Temporal-order interventions preserve the same 32 real frames and modify only
their ordering. As shown in Table~\ref{tab:evidence}(b), R(2+1)D degrades under
both shuffling and reversal, while MViT shows smaller but consistent drops.
Video Swin-T exhibits no stable degradation under either intervention, revealing
substantial architecture-dependent differences in temporal-order sensitivity. The evaluated ViViT is an adapted factorized baseline rather than a complete
reproduction of the original architecture. It uses no temporal positional encoding
and applies temporal mean pooling, making it inherently insensitive to frame
permutations; accordingly, neither shuffling nor reversal changes its AUC. Overall, video input does not guarantee temporal-order utilization, and temporal
sensitivity alone does not predict Cross-Env robustness.

\begin{figure}[t]
\centering
\includegraphics[width=0.84\linewidth]{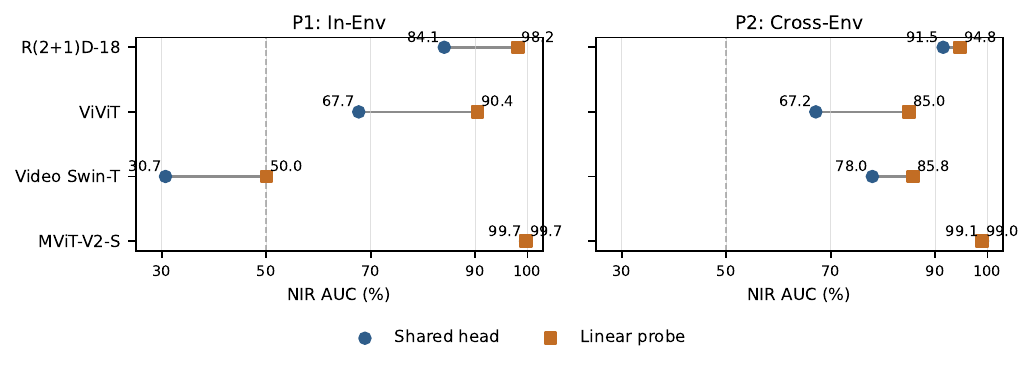}
\vspace{-2.0em}
\caption{Frozen-backbone NIR diagnosis. The shared head uses the original multimodal classifier, while the linear probe is trained on paired-validation NIR embeddings and evaluated on paired test samples.}
\label{fig:nirprobe}
\end{figure}

\noindent\textbf{NIR Representation Diagnosis.}
NIR-only performance differences may originate from different mechanisms. To distinguish classifier-head misalignment from limited linear separability of single-stream representations, we compare the original shared classifier with a frozen-backbone linear probe. Figure~\ref{fig:nirprobe} illustrates this diagnosis: the shared head directly uses the original multimodal classifier, whereas the linear probe freezes the NIR backbone and trains a new linear classifier on paired-validation NIR embeddings before evaluation on paired test samples. Because P1 and P2 use independently trained checkpoints and different test partitions, we interpret the probe as a \emph{within-protocol} recoverability analysis rather than a direct comparison of absolute NIR difficulty across protocols. MViT remains near 99\% AUC with both the shared head and linear probe, indicating that its original classifier already exploits the available NIR representation effectively. R(2+1)D improves from 84.09\% to 98.17\% on P1, while ViViT improves from 67.71/67.14\% to 90.42/85.01\% on P1/P2, indicating that their NIR representations retain substantial linearly accessible information that is not fully utilized by the original shared classifier. In contrast, Swin reaches only 50.00\% AUC with the P1 linear probe, indicating weaker linear separability of its single-stream NIR representation under this diagnostic. Thus, NIR-only degradation has no single cause across architectures, but can arise from either classifier adaptation limitations or restricted linear separability of the learned representation.

\vspace{-1.0em}
\section{Conclusion}
\vspace{-1.0em}

We introduced \textbf{GBU-Palm}, a large-scale multimodal video dataset and
benchmark for palm presentation attack detection. Systematic evaluation shows
architecture-dependent sensitivity to environmental shift, non-uniform RGB-NIR
complementarity, and substantially different reliance on temporal order.
Decision-conditioned analysis and NIR probing further reveal that similar aggregate
performance can hide different security and usability failures, while single-modality
degradation may arise from either classifier- or representation-related limitations.
Overall, benchmark performance, information utilization, and cross-environment
transfer are related but distinct properties of palm PAD models. Future extensions
will include three-dimensional and more challenging attacks, additional sensing
modalities, and richer spatiotemporal annotations.
\bibliographystyle{splncs04}
\bibliography{references}

@article{bhilare2018vulnerability,
  author={Shruti Bhilare and Vivek Kanhangad and Narendra S. Chaudhari},
  title={A Study on Vulnerability and Presentation Attack Detection in Palmprint Verification System},
  journal={Pattern Analysis and Applications}, volume={21}, number={3}, pages={769--782}, year={2018},
  doi={10.1007/s10044-017-0606-y}
}

@inproceedings{tome2015vulnerability,
  author={Pedro Tome and S{\'e}bastien Marcel},
  title={On the Vulnerability of Palm Vein Recognition to Spoofing Attacks},
  booktitle={International Conference on Biometrics (ICB)}, pages={319--325}, year={2015},
  doi={10.1109/ICB.2015.7139056}
}

@inproceedings{yao2023palmprint,
  author={Dingyi Yao and Huikai Shao and Dexing Zhong},
  title={Palmprint Anti-Spoofing Based on Domain-Adversarial Training and Online Triplet Mining},
  booktitle={IEEE International Conference on Image Processing (ICIP)}, pages={1235--1239}, year={2023},
  doi={10.1109/ICIP49359.2023.10223182}
}

@article{liu2025domain,
  author={Chengcheng Liu and Huikai Shao and Dexing Zhong},
  title={Learning Domain-Adaptive Palmprint Anti-Spoofing Feature from Multi-Source Domains},
  journal={Displays}, volume={86}, pages={102871}, year={2025},
  doi={10.1016/j.displa.2024.102871}
}

@article{liu2025hfsra,
  author={Chengcheng Liu and Huikai Shao and Dexing Zhong},
  title={Learning Discriminative Palmprint Anti-Spoofing Features via High-Frequency Spoofing Regions Adaptation},
  journal={IET Image Processing}, volume={19}, number={1}, year={2025},
  doi={10.1049/ipr2.70029}
}

@article{yan2026pvasd,
  author={Caiping Yan and Zhi Lan and Hong Li and Yuqi Li and Zonglin Meng},
  title={A Comprehensive Framework for Palm Vein Anti-Spoofing With Preprocessing Pipeline, Dataset, and Benchmark},
  journal={IEEE Transactions on Information Forensics and Security}, volume={21}, pages={945--959}, year={2026},
  doi={10.1109/TIFS.2025.3650391}
}

@inproceedings{liu2018deepfas,
  author={Yaojie Liu and Amin Jourabloo and Xiaoming Liu},
  title={Learning Deep Models for Face Anti-Spoofing: Binary or Auxiliary Supervision},
  booktitle={IEEE/CVF Conference on Computer Vision and Pattern Recognition (CVPR)}, pages={389--398}, year={2018},
  doi={10.1109/CVPR.2018.00048}
}

@inproceedings{tran2018r2plus1d,
  author={Du Tran and Heng Wang and Lorenzo Torresani and Jamie Ray and Yann LeCun and Manohar Paluri},
  title={A Closer Look at Spatiotemporal Convolutions for Action Recognition},
  booktitle={IEEE/CVF Conference on Computer Vision and Pattern Recognition (CVPR)}, pages={6450--6459}, year={2018},
  doi={10.1109/CVPR.2018.00675}
}

@inproceedings{arnab2021vivit,
  author={Anurag Arnab and Mostafa Dehghani and Georg Heigold and Chen Sun and Mario Lu{\v{c}}i{\'c} and Cordelia Schmid},
  title={ViViT: A Video Vision Transformer},
  booktitle={IEEE/CVF International Conference on Computer Vision (ICCV)}, pages={6836--6846}, year={2021}
}

@inproceedings{liu2022videoswin,
  author={Ze Liu and Jia Ning and Yue Cao and Yixuan Wei and Zheng Zhang and Stephen Lin and Han Hu},
  title={Video Swin Transformer},
  booktitle={IEEE/CVF Conference on Computer Vision and Pattern Recognition (CVPR)}, pages={3202--3211}, year={2022}
}

@inproceedings{li2022mvitv2,
  author={Yanghao Li and Chao-Yuan Wu and Haoqi Fan and Karttikeya Mangalam and Bo Xiong and Jitendra Malik and Christoph Feichtenhofer},
  title={MViTv2: Improved Multiscale Vision Transformers for Classification and Detection},
  booktitle={IEEE/CVF Conference on Computer Vision and Pattern Recognition (CVPR)}, pages={4804--4814}, year={2022}
}

@inproceedings{zhang2019casiasurf,
  author={Shifeng Zhang and Xiaobo Wang and Ajian Liu and Chenxu Zhao and Jun Wan and Sergio Escalera and Hailin Shi and Zezheng Wang and Stan Z. Li},
  title={A Dataset and Benchmark for Large-Scale Multi-Modal Face Anti-Spoofing},
  booktitle={IEEE/CVF Conference on Computer Vision and Pattern Recognition (CVPR)},
  pages={919--928}, year={2019}
}

@inproceedings{george2021crossmodal,
  author={Anjith George and S{\'e}bastien Marcel},
  title={Cross Modal Focal Loss for RGBD Face Anti-Spoofing},
  booktitle={IEEE/CVF Conference on Computer Vision and Pattern Recognition (CVPR)},
  pages={7882--7891}, year={2021}
}

@inproceedings{yang2019modeldata,
  author={Xiao Yang and Wenhan Luo and Linchao Bao and Yuan Gao and Dihong Gong and Shibao Zheng and Zhifeng Li and Wei Liu},
  title={Face Anti-Spoofing: Model Matters, so Does Data},
  booktitle={IEEE/CVF Conference on Computer Vision and Pattern Recognition (CVPR)},
  pages={3507--3516}, year={2019}
}
\end{document}